%% file: ms.tex
\documentclass[10pt,twocolumn,letterpaper]{article}

\usepackage{wacv}
\usepackage{times}
\usepackage{epsfig}
\usepackage{graphicx}
\usepackage{amsmath}
\usepackage[dvipsnames]{xcolor}
\usepackage{amssymb}
\usepackage{booktabs}

\DeclareMathOperator*{\argmin}{arg\,min}

\def\wacvPaperID{1472} 

\wacvalgorithmstrack   

\wacvfinalcopy 

\ifwacvfinal
\usepackage[breaklinks=true,bookmarks=false,hidelinks]{hyperref}
\else
\usepackage[pagebackref=true,breaklinks=true,bookmarks=false,hidelinks]{hyperref}
\fi
\hypersetup{
    colorlinks=true,
    linkcolor=blue,
    filecolor=magenta,      
    urlcolor=cyan,
}
\newcommand{\ours}{\textit{SPAN}\xspace}

\begin{document}
\hypersetup{urlcolor = [rgb]{0,0,0.5}, citecolor = [rgb]{0,0,0.5}}

\title{Prior Knowledge-Guided Attention in Self-Supervised Vision Transformers}

\author{Kevin Miao, Akash Gokul, Suzanne Petryk, Raghav Singh, Joseph Gonzalez, Kurt Keutzer, \\ Trevor Darrell, Colorado Reed
\\ \\ UC Berkeley\\
{\tt\small \{kevinmiao, akashgokul, spetryk, raghavsingh, jegonzal, keutzer, trevor, cjrd\}@berkeley.edu}
}

\maketitle
\thispagestyle{empty}

\input{0-Abstract}


\input{1-Introduction}

\input{2-Related}

\input{3-Methods}

\input{4-Results}

\input{5-Conclusion}

{\small
\bibliographystyle{ieee_fullname}
\bibliography{egbib}
}

\input{6-Appendix}

\end{document}

%% file: 0-Abstract.tex
\begin{abstract}
Recent trends in self-supervised representation learning have focused on removing inductive biases from training pipelines. However, inductive biases can be useful in settings when limited data are available or provide additional insight into the underlying data distribution. We present spatial prior attention (\ours), a framework that takes advantage of consistent spatial and semantic structure in unlabeled image datasets to guide Vision Transformer attention. \ours operates by regularizing attention masks from separate transformer heads to follow various priors over semantic regions. These priors can be derived from data statistics or a single labeled sample provided by a domain expert. We study \ours through several detailed real-world scenarios, including medical image analysis and visual quality assurance. We find that the resulting attention masks are more interpretable than those derived from domain-agnostic pretraining. \ours produces a 58.7 mAP improvement for lung and heart segmentation. We also find that our method yields a 2.2 mAUC improvement compared to domain-agnostic pretraining when transferring the pretrained model to a downstream chest disease classification task. Lastly, we show that \ours pretraining leads to higher downstream classification performance in low-data regimes compared to domain-agnostic pretraining.
\end{abstract}

%% file: 1-Introduction.tex
\section{Introduction}

\begin{figure*}[!t]
    \hspace{-1.5em}
    \centering
\includegraphics[width=1.0\textwidth]{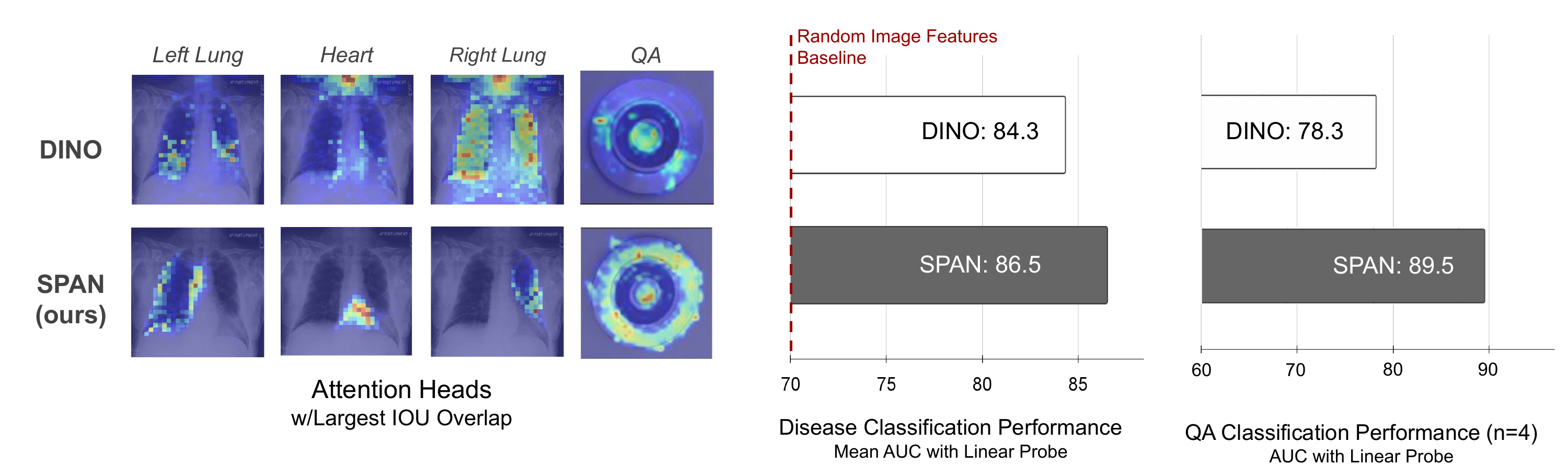}
    \caption{\textbf{Self-attention from a Vision Transformer on chest X-rays and QA data, where the attentions heads with the largest IOU overlap with the region of interest.} Existing self-supervised training methods for Vision Transformers, such as DINO, learn scattered attention maps that do not necessarily attend to the constituent objects within the image. \ours, on the other hand, uses prior knowledge to guide the attention to such regions, as visible in the maps. As indicated in the bar plot on the right, constraining the attention to these semantic components leads to better performing representations as determined by a linear probe, multi-label disease classification (CheXpert) \cite{irvin2019chexpert} and low-data QA tasks \cite{test} -- see Section \ref{sec:exps} for details.}
    \label{fig:teaser}
    \vspace{-10pt}
\end{figure*}

Recent works in unsupervised learning have largely focused on removing inductive biases from the training process: transformer-based methods have successfully removed the scale-and-shift invariance from CNNs~\cite{dosovitskiy2020image} and autoencoders have successfully removed the hardcoded augmentation-based invariances from contrastive learning methods~\cite{he2021masked}. 
 However, inductive biases can capture knowledge that would otherwise be difficult to infer strictly from observed data and are particularly beneficial when there is not enough data to generalize toward unseen scenarios~\cite{baxter2000model,bouvier2020robust,murphy2012machine}. Another instance where inductive biases can may useful is when domain-specific knowledge provides additional information that can be used to model the underlying data distribution~\cite{raissi2019physics,desai2021parsimonious}.


Medical imaging and quality assurance in manufacturing (QA) are two important, real-world applications that can benefit from the inclusion of inductive biases. Medical imaging is an application where data are expensive to acquire and store, require rigorous labeling procedures by certified experts, and are subjected to a plethora of privacy and regulatory concerns\cite{kelly2019key}. Manufacturing data are often considered private and the development of large datasets is therefore difficult. Additionally, the complex nature of the problem due to recognition of extremely fine-grained details as well as imbalanced datasets make it difficult to train highly accurate QA models \cite{lucas2019}. However, a benefit of both these applications is that individual data points share a common underlying structure that is well understood. This underlying structure can provide a strong inductive bias that a model might not be able to learn from data alone or accelerate training with fewer data points. For instance, the human anatomy informs about the presence and relative positioning of organs and the type of image (e.g. X-ray, CT, MRI) reveals information about the characteristics of the pixel intensities. In manufacturing QA, the same holds for a blueprint, mould or model template with respect to individual items.

While one thread of research seeks to remove such inductive biases and learn directly from data, another thread seeks to include useful inductive biases to guide the training process, particularly in structured domains like medical imaging and manufacturing QA. In this work, we investigate an intersection of these two threads where we leverage both a self-supervised Vision Transformer learning framework as well as spatial priors to guide the training.

We present a framework called spatial prior attention (\ours) that builds upon self-supervised Vision Transformers \cite{caron2021emerging} and leads to more interpretable attention heads, higher downstream classification task performance as well as performance improvements when little annotated data are available -- see Figure~\ref{fig:teaser}. \ours incorporates prior knowledge into self-supervised training of Vision Transformers by regularizing a subset of attention heads in the multi-headed self-attention module to attend to objects of interest. Instead of relying on human annotations of these images to determine the boundaries of these objects, \ours leverages the shared underlying structure to define a template of object boundaries (e.g. individual organs or parts). It then aligns and registers each image to this template, transforms the object boundaries accordingly, and then uses these obtained object boundaries to regularize the attention heads throughout the training procedure -- Figure~\ref{fig:medino} provides an overview of this process which is detailed in Section~\ref{sec:methods}.

Compared to DINO \cite{caron2021emerging}, a recent self-supervised Vision Transformer framework, \ours leads to better downstream task performance, and more interpretable attention maps. Specifically, \ours leads to more semantically meaningful attention maps with a 58.7 mAP as measured by their overlap with ground truth lung and heart segmentation. This also yields 2.2 mAUC improvement in the CheXpert disease classification task. Additionally, in the presence of a handful of training examples, \ours pretrained models show higher quality assurance classification results as compared to domain-agnostic pretrained models. To summarize, we present the following contributions:

\begin{enumerate}
\item We use self-supervised Vision Transformers to present a novel knowledge-guided attention regularization framework that leverages attention modules to learn disentangled and meaningful representations.
\item We establish a range of procedures that incorporate prior knowledge and inductive biases into templates when annotated data are sparse or additional priors reveal information that cannot be learned from the data alone\cite{rieger2020interpretations}.
\item We find that encoding prior knowledge using attention regularization yields more semantic meaningful attention maps. Specifically, the overlap between the learned attention maps and ground truth lung and heart segmentations improves by 58.7 mAP, compared to domain-agnostic pretraining. This leads to a 2.2 mAUC increase in the downstream CheXpert disease classification tasks.
\item We show that in the presence of only a handful of examples, \ours pretrained models attain improvement gains over the baseline in manufacturing defects detection. Provided with 4 training images, \ours pretrained models see a 11.2 increase in AUC over domain-agnostic pretrained models.
\end{enumerate}

%% file: 2-Related.tex
\section{Related Works}

\ours incorporates domain knowledge to improve the performance and interpretability of self-supervised pretraining for medical images. It builds upon works in self-supervised learning, knowledge-guided and interpretable methods, and image registration and alignment, which we detail below.

\textbf{Self-Supervised Learning.}
The performance of machine learning models is heavily contingent on the choice of features and representations from which they learn. Representation learning aims to reveal these intrinsic qualities of data such that they are informative and effective for a desired task \cite{bengio2013representation}, such as image classification or object detection. Contemporary methods involve contrastive learning based approaches \cite{he2020momentum,chen2020simple}, clustering-based techniques \cite{caron2020unsupervised}, and self-distillation \cite{grill2020bootstrap,caron2021emerging}. DINO \cite{caron2021emerging} is an example of a self-supervised learning framework that uses self-distillation, yielding state-of-the-art downstream performance using the Vision Transformer (ViT) architecture \cite{dosovitskiy2020image}. We focus on this framework for two reasons: (i) the attention modules in ViT allow for greater interpretability than CNN-based approaches that require external tools such as Grad-CAM~\cite{wang2021gradcam} to extract pixel-level saliency relationships, (ii) in self-supervised training, the DINO attention maps have the demonstrated ability to segment salient foreground objects~\cite{caron2021emerging}, which, as we show, provide a strong mechanism to regularize salient objects in \ours.

Most existing self-supervised algorithms are benchmarked based on their performance following pretraining on generic, object-centric image datasets, such as ImageNet, which potentially leads to poor results in domains where the data are dissimilar to these datasets~\cite{xiao2020should,reed2022self, Reed_2021_CVPR}. Furthermore, domain-specific applications of machine learning can benefit from self-supervised pretraining due to the cost and expertise needed to accurately annotate data \cite{sowrirajan2021moco,gazda2021self,bai2019self,azizi2021big}.
For instance, MoCo-CXR \cite{sowrirajan2021moco} adapts MoCo \cite{he2020momentum} pretraining to chest X-ray data by designing new data augmentations suitable for recognizing subtle differences between X-ray images. We use MoCo-CXR's data augmentations as it uses similar X-ray images as our work. IDEAL \cite{mahapatra2021interpretability} focuses on self-supervision and interpretability. However, they use saliency reconstruction to find informative samples for active learning and do not focus purely on learning discriminative and interpretable representations. 


\textbf{Knowledge-Guided Learning.}
Knowledge-guided learning seeks to incorporate prior knowledge in such a way that it leads to better performance, efficiency, or interpretability for the learned model. Several papers have incorporated first order logic rules~\cite{hu2020logicweights,roychowdhury2021logicconstraints,yin2019domain} as well as anatomical constraints for pose estimation~\cite{ning2017knowledge,bigalke2021domain}.  
Neural networks have also been combined with physics-based models to capture and enforce the relationship between variables through an additional loss~\cite{daw2017physics}, which is similar to our work in that it also captures the alignment between the guided attention mechanism and domain knowledge via an attention-based loss mechanism.
Different from these works, \ours is an attention-based regularization in transformer models to enforce such constraints; \ours can be easily added to existing models without the need for making large changes. Furthermore, attention based approaches have been used to improve the explainability of computer vision models through visualizations of attention maps to indicate important regions~\cite{xu2016satcaption}. Convolutional neural networks use tools such as CAM \cite{zhou2015cnnlocalization} and GradCAM \cite{wang2021gradcam} to create attention maps by looking at the hidden layer activations. Another approach is the Attention Branch Network \cite{fukai2018abn} that generates an attention map based on the extracted features and then uses it to mask out irrelevant features. These attention maps are evaluated through visual checks or against segmentation datasets which are limited in the medical domain. As discussed in Section~\ref{sec:methods}, our paper instead uses image registration to align the attention maps with an inductive bias corresponding to a salient region so only a single representative sample is required.

\textbf{Image Registration.} Image registration is the task of projecting one image onto the coordinate system of another image with similar content \cite{medimreg}. This classical challenge has been devised for the medical domain as it facilitates the development of atlases. In this paper, we are particularly interested in using a specific type of image registration: deformable image registration. This is useful as differences in individual specimens can be modeled as deformable transformation to an exemplar, i.e. morphological differences in organs in different humans or manufacturing discrepancies \cite{sotiras2013deformable}. Traditionally popular techniques to solve this alignment problem involve congealing\cite{cox2008least} or optical flow \cite{Lefbure2001}. Here we focus on b-Spline registration due to its simplicity and efficiency, as it only requires one data example. It operates by modeling the deformation field as b-spline curves where each pixel maps each voxel in the source image to the target image \cite{Shackleford2013}. We favored this method over state-of-the-art neural methods\cite{2019vmorph,mok2020fast,sloan2018learning}, as these require more compute and data points. Mansilla et. al. \cite{mansilla} embed prior knowledge in the form of anatomical constraints to improve image registration tasks. Our work differs drastically as we aim to improve self-supervised pretraining methods using deformable transformations in both medical and non-medical applications.

%% file: 3-Methods.tex
\section{Spatial Prior Attention}
\label{sec:methods}

\begin{figure*}[!t]
    \begin{center}
    \hspace{-1.5em}
    \centering
\includegraphics[width=0.8\textwidth]{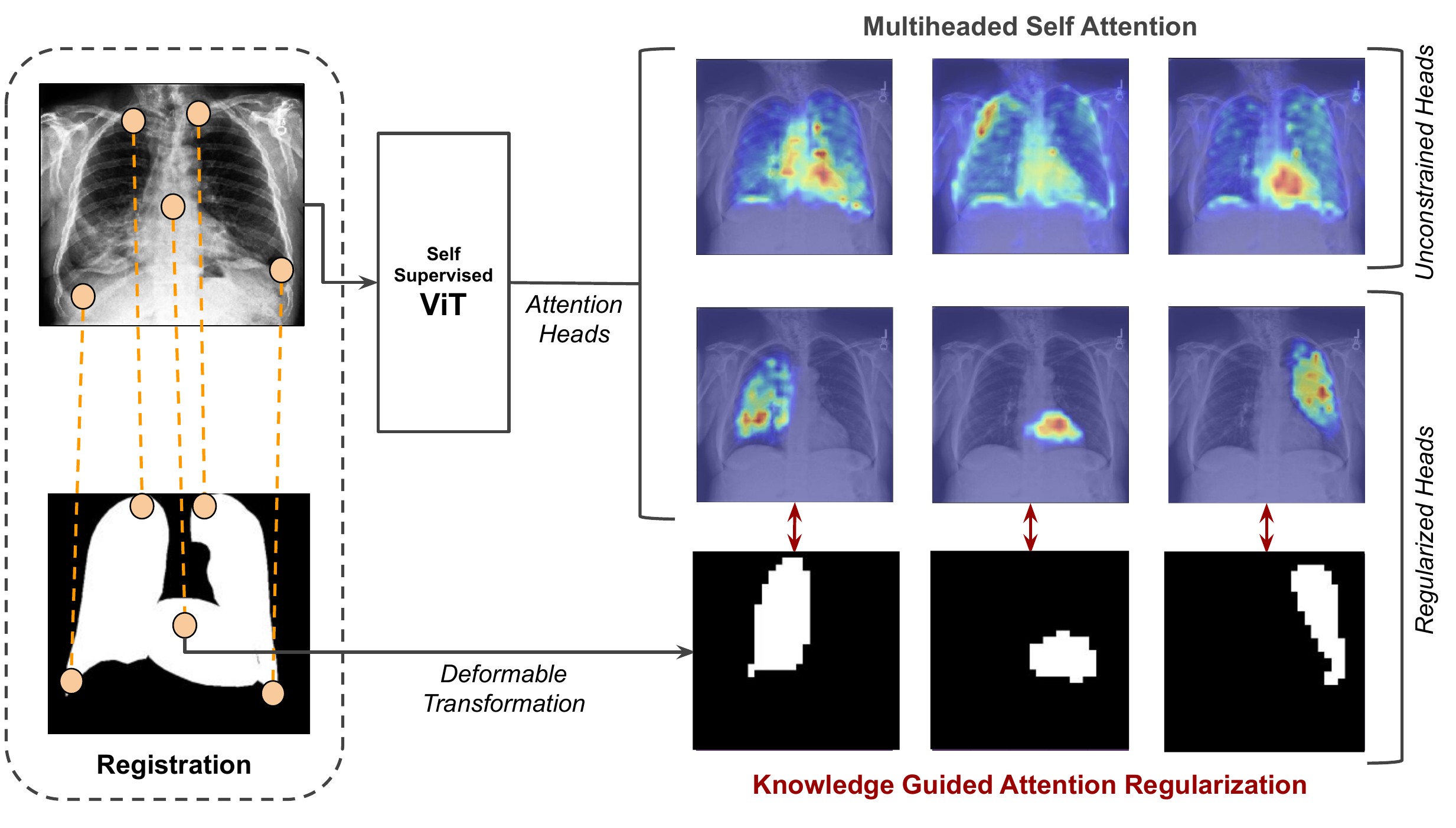}
    \caption{\textbf{The \ours framework}. \ours first registers each image to an exemplar template with known segmentations, and the registration outputs a deformable transformation that is applied to the template. During self-supervised pretraining with a ViT model, each component of the template then regularizes an individual attention head in the multiheaded self-attention modules (Regularized Heads). A subset of the attention heads are also unconstrained (Unconstrained Heads). }
    \label{fig:medino}
    \vspace{-15pt}
    \end{center}
\end{figure*}

The goal of \ours is to incorporate domain knowledge to improve both the performance and quality of representations learned through self-supervised pretraining.
To do so, \ours regularizes transformer attention heads to follow inductive biases on a shared common semantic structure among instances in the dataset, as seen in Figure~\ref{fig:medino}. For example, we can incorporate the inductive bias that chest radiographs have expected anatomical relationships between the relative positions of the lungs and heart. In the following sections, we detail a means of effectively incorporating this type of knowledge in the form of simple spatial heuristics or even a single instance of ground truth knowledge, into the DINO pretraining of Vision Transformers. 

\subsection{Self-Supervised Vision Transformers with Knowledge Distillation}

Caron et al. ~\cite{caron2021emerging} present a transformer-based knowledge distillation technique, DINO, that we build upon for \ours. In DINO, a student model $m_{\theta_{s}}$ is trained to match the output of a teacher model $m_{\theta_{t}}$ (parameterized by $\theta_{s}$ and $\theta_{t}$ respectively). This distillation objective is reframed as a representation learning objective where representations are learned for each of $n$ different views of original image $X$, $\{X_1, ..., X_n\}$, obtained via a set of data augmentations $V$. The DINO objective encourages the student model to learn ``local-to-global" correspondences. This happens by passing in local and global crops of an image to the student and tasking the student model to predict the teacher's representation. The teacher is only given global crops denoted $X^{g}_1$ and $X^{g}_2$. To train the student network, DINO defines the probability distributions $P_m$ for the student and teacher model:
$$P_m(X) = softmax\left(\dfrac{m_{\theta_{m}}(X)}{\tau_{m}}\right)$$
where $\tau_m$ is the model-specific temperature. The overall DINO objective, given below, is the cross-entropy loss $H(p,q) = -p\log q$ over the probability distributions $P_s(X)$ and $P_t(X)$. 

$$ L(X_1, X_2) = H(P_{t}(X_1), P_{s}(X_2))$$
$$ = \argmin_{\theta_{s}} \sum_{X \in \{X^{g}_1, X^{g}_2\}} \sum_{X^{\prime} \in V} L(X, X')$$

While the original DINO augmentations can be powerful for learning representations from a dataset such as ImageNet, they can fail in domains where local structure is critical to scene understanding\cite{xie2021unsupervised}. For instance, local crops can be harmful when applied to multi-object image datasets as compared to object-centric data because different views potentially correspond to completely different objects.

Our preliminary empirical findings (see 4.3 and 4.4) corroborate this as local crops harm the performance of self-supervised learning on the datasets studied in this paper. Following this, we instead use a set of domain-specific augmentations which replace DINO's local crops with other task-relevant data augmentations (see the Appendix A2). 

\subsubsection{Attention}

Vision Transformers perform well on a wide variety of vision tasks and allow for pixel-level relationship introspection due to their built-in attention modules \cite{dosovitskiy2020image}. As input, Vision Transformers take in a sequence of $P$ image patches with fixed size ($p=16$) which is prepended by a $[CLS]$-token. The $[CLS]$ token enables a corresponding output that allows for downstream tasks such as classification.

Self-attention modules are a key component in Transformer networks. Given embeddings $q, k, v$ calculated from a sequence of inputs, the attention matrix $A$ measures the pairwise similarity between $q_i$, query value of patch $i$, in relationship with $k_j$, key value of patch $j$. Formally, 
$$A = \sigma{\left(\dfrac{qk^{\intercal}}{\sqrt{D_{h}}}\right)}$$
where $D_h$ is defined as the dimensionality of the heads and $A \in \mathbb{R}^{P \times P}$. When probing self-attention, we extract the attention values of each patch with respect to the $[CLS]$ token of the last layer of each of $n_h$ heads and exclude the attention value for the $[CLS]$ token with itself. This tensor is then upsampled via nearest-neighbor interpolation into the shape of the original image resulting in an attention map $A_{s} \in \mathbb{R}^{w \times h \times n_h}$ where $w$ and $h$ are the dimensions of $X$.

\subsection{Knowledge-Guided Regularization}

The Vision Transformer's attention module allows us to guide a model given any arbitrary binary knowledge map $K$ by back-propagating through the model. If $K \in \{0, 1\}^{w \times h}$, where $K_{ij}$ is $1$ if the patch at location $i, j$ is considered a useful bias and $0$ otherwise, we incur an exclusion loss for models' self-attention map $A_{s}^{(\theta_t)}$ attending outside salient regions and a negative inclusion loss for attending at salient regions. This yields the following regularization terms that are combined with the DINO objective:
\begin{eqnarray*}
L_{inclusion}(A, K) = {\lambda_1} \sum_{i}  \sum_{j} a_{ij}^{(\theta_t)} k_{ij} \\
L_{exclusion}(A, K) = - {\lambda_2} \sum_{i} \sum_{j} a_{ij}^{(\theta_t)} (1 - k_{ij})
\end{eqnarray*}


where hyperparameters $\lambda_1, \lambda_2$ denote the respective regularization strengths. Experimentally, we find that both regularizers are needed in order to prevent mode collapse in the attention maps (See Appendix A4).

\subsubsection{Attention Head Specialization}

Vision transformers have multiple independent attention heads. These attention heads $A$ have the capability to attend to different sub-modular entities present within an input $X$\cite{vaswani2017attention}. When prior knowledge underlying a certain task postulates a common underlying semantic and spatial structure, these attention heads can be instructed to attend to specific elements. Specifically, this knowledge on structures expected to be present in an attention head $A_{i}$ can be embedded through a binary knowledge map $K_{i} \in \{0, 1\}^{w \times h}$ for up to $n$ attention heads. If the number of known entities are lower than the total number of attention heads, unassigned heads remain unregularized and attend to general regions.

The incorporation of this information enables task-specificity and functions as a scaffold for interpretability through which failure cases can be deconstructed into explainable task sub-entities. Knowledge maps representing a specific task can be assigned arbitrarily to any attention head. 

\subsection{Encoding Knowledge into Templates}

\begin{figure}[!t]
    \begin{center}
    \hspace{-1.5em}
    \centering
\includegraphics[width=1.0\linewidth]{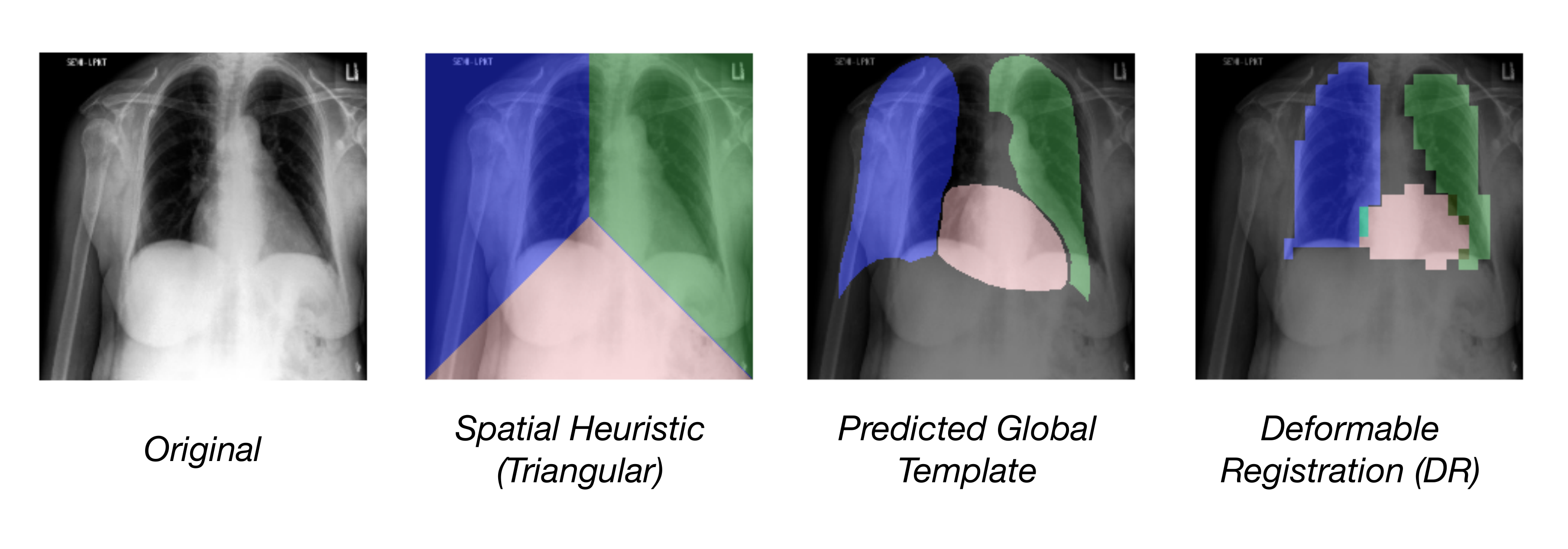}
    \caption{\textbf{Example templates for encoding spatial and semantic information.} 1st image: a randomly sampled image from the CheXpert dataset. 2nd image: a template based on spatial heuristics. 3rd image: a global prediction-based template. These masks are computed by averaging the predictions made from an external segmentation model. 4th image: deformable registration template. Given an exemplar image with ground truth segmentation mask, the template is obtained by warping the segmentation using deformable image registration.}
    \label{fig:KnowledgeTemplates}
    \vspace{-20pt}
    \end{center}
\end{figure}

Our regularization procedure allows for any type of inductive bias to be embedded that can be translated into knowledge map $K$, as described above. We identified two types of inductive biases that are useful in guiding vision models where the input data is decomposable into multiple elements: (1) spatial and (2) semantic. These categories are then used to embed prior knowledge, such as anatomical constraints or other assumptions, into a knowledge map $K$. Intuitively, the goal is to not only assist a model to look at task-relevant features but also to specialize the individual heads. To this end, we assign specific heads $n \in N$ to discrete elements of interest. The remainder of the attention heads remain unassigned and hence, are able to attend the whole image $X$ without any restrictions. We use the following three knowledge encoding procedures for \ours which are depicted in Figure \ref{fig:KnowledgeTemplates}:

\paragraph{Spatial Heuristic.} 
As a baseline, we explore a simple spatial heuristic that approximately segments the constant relative positioning of organs in the thorax into a knowledge map $K$. We encode our knowledge as a tripartite mask with triangular parts corresponding to the left lung, right lung, and the heart. 

\paragraph{Predicted Global Template.} Instead of relying on nonspecific spatial regions for attention supervision, we calculate a global average of the predicted locations of relevant organs from a pretrained model. We trained a segmentation model (DeepLabv3-ResNet101 \cite{chen2017rethinking}) on 200 chest X-rays with labeled heart and lung segmentation data \cite{shiraishi2000development}. We average the model's inference segmentations over all images in the pretraining image dataset to obtain a single predicted global template. These knowledge maps provide a more robust spatial bias signal than the spatial heuristic.

\paragraph{Semantic Deformable Image Registration.} 
To test the impact of increasingly accurate knowledge templates, we use a single ground truth segmentation from a different dataset which is adapted to our dataset via deformable image registration. Given a single annotated exemplar pair of image $X^{e}$ and its ground truth segmentation $S^{e}$, canonical deformable image registration \cite{sotiras2013deformable} is performed to learn a parameterization $\phi$ that deforms exemplar image $X^{e}$ to training image $X^{i}$. This learned $\phi$ is then used to create a deformable knowledge map $K$, as an estimate to true $S_{i}$, by applying $\phi$ on $S_{e}$. In our paper, we use \textit{SimpleElastix} wrappers \cite{lowekamp2013design,yaniv2018simpleitk,beare2018image} that are based on b-Spline deformation models. As seen in Figure~\ref{fig:KnowledgeTemplates}, this procedure results in the most accurate results due to combining both spatial and semantic information. 


%% file: 4-Results.tex
\section{Experiments and Results}
\label{sec:exps}

 In the following experiments, we compare the qualitative and quantitative performance of \ours with self-supervised Vision Transformers without attention regularization. The different experiments focus on interpretability and downstream classification performance. The quantitative and qualitative interpretability analyses reveal that \ours leads to more interpretable representations, the second set of experiments show the increased downstream classification performance, and the third set shows the data-efficiency of \ours on quality assurance manufacturing data. 
 
\subsection{Setup}
\paragraph{Datasets}

We pretrain our models using CheXpert, a medical X-ray dataset with 190k images and 14 disease classes collected from 65,240 unique patients \cite{irvin2019chexpert}. We exclude the lateral images, as no high-quality lateral image priors are available. To measure the accuracy of the learned representation, we evaluate them using mean average precision scores (mAP) to quantify the overlap of learned attention with semantic regions against two ground truth segmentation datasets, JSRT \cite{shiraishi2000development} and Montgomery \cite{jaeger2014pu,rajaraman2021improved}. These two datasets are smaller in size and contain 247 and 138 images respectively. JSRT provides segmentation masks for both lungs and the heart. Montgomery only has annotations for the lungs. Lastly, we also use a QA manufacturing dataset extracted from Kaggle \cite{test}. This set consists of 1,300 images of molds with 781 defect and 519 functioning objects. The goal of this challenge is to create models that can classify whether an entity is defect or functioning.

\paragraph{Data Augmentations}
In our experiments, we differentiate between domain-agnostic and domain-specific data augmentations. Domain-agnostic data augmentations are based on the default DINO \cite{caron2021emerging} and BYOL \cite{grill2020bootstrap} augmentations. They contain global crops, local crops, color jittering, Gaussian blur and solarization. \ours incorporates domain-specific data augmentations, in particular chest X-ray specific data augmentations, are inspired from ChX-MoCo \cite{sowrirajan2021moco}, a framework for Momentum Contrasting in X-rays. These augmentations perform global crops (not local crops) in addition to translations, rotations, brightness, contrast and sharpness.

\paragraph{Training and Finetuning}

In our experiments, we fixed the backbone of our models to be the small Vision Transformer (ViT-S, 21M parameters) with patch size 16. Self-supervised pretraining is performed on 8 GPUs (NVIDIA Tesla V100). We train Imagenet pretrained (800 epochs) ViTs using an Adam optimizer, batch size $28$, and a base learning rate of $10^{-3}$ for $30$ epochs. Other hyperparameters are directly implemented from DINO\cite{caron2021emerging}. The best attention regularization hyperparameters $\lambda_1$ and $\lambda_2$ are chosen using a sweep for values between $[10^{-2}, 10^{-6}]$. For downstream classification tasks, we train a linear layer on top of the frozen learned representations without any sort of data augmentations for 100 epochs.

\input{tables/fig-A}

\subsection{Attention Head Interpretability}

\begin{figure*}[!t]
\vspace{-5pt}
    \begin{center}
    \hspace{-1.5em}
    \centering
\includegraphics[width=0.80\textwidth]{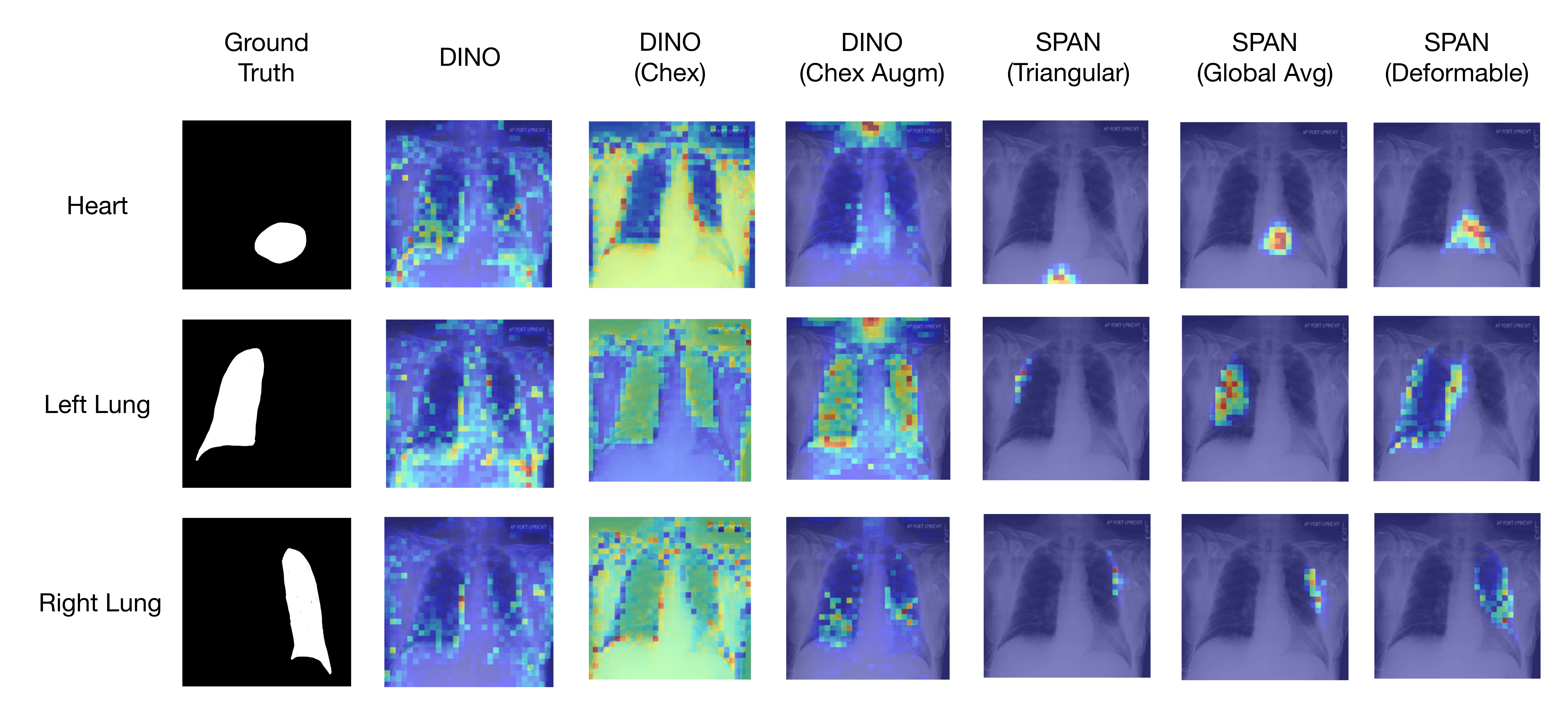}
    \caption{\textbf{Visualized attention maps resulting from different pretraining procedures.} We analyze the visualized attention maps by probing the heads of the respective differently pretrained models and. We choose the map with the highest IOU overlap with the ground truth for each model. These maps show that as the prior for attention becomes more specific, the mAP and specialization of attention heads increases. Additionally, DINO models are unable to learn interpretable representations without chest-specific augmentations.}
    \label{fig:MEPrune}
    \vspace{-20pt}
    \end{center}
\end{figure*}

\subsubsection{Performance}
The goal of our evaluation is to evaluate the quality of the learned representations. Specifically, we are interested in the interpretability aspect. In this paper interpretability is defined as the ability to understand, debug and justify the model's decision processes \cite{molnar2020interpretable}. Explainable representations are those of which the attention can be used to attend to meaningful regions of an image - i.e. demarcated regions \cite{selvaraju2017grad} \cite{liu2021visual}.

In Table 1, we compare the different models' attention maps against the ground truth segmentations for the lungs and heart. The Montgomery dataset does not contain ground truth segmentation maps for the heart and hence these results have been omitted. The interpretability results are evaluated using pixel-wise mAP scores that calculate the average precision at different thresholds. For the \ours trained models, we use the attention maps at the assigned head for evaluation. In DINO tasks where no head was assigned to a specific part, the score represents the maximum across the different heads. 

The interpretability results show higher mAP scores in \ours models compared to DINO pretrained models for all thoracic parts. \ours with triangular spatial heuristics sees a 30 mAP increase in performance over the DINO baseline pretrained using chest specific augmentations. This further improves with the templates acquired from global average masks, specifically in the right lung. As the templates become more specific, the deformable semantic masks acquired further performance gains yielding 88.5 mAP on average in JSRT. This is a 58.7 mAP increase over the baseline (DINO with chest augmentations). \textit{The key trend we observe is that the more specific information that is encoded in masks, the higher the interpretability scores}. The results also corroborate that semantic and spatial information ultimately attain the highest performance outcomes, as the deformable, parts-based mask model attained the highest performance. In general, any attention based model seems to outperform a non-guided model.

The baseline DINO pretrained with domain-agnostic augmentations on X-ray scans has the lowest scores across all body parts. 
Interestingly, a pretrained model that was not pretrained on chest images outperforms this setup. 
This suggests that DINO domain-agnostic augmentations (such as the global and local crops) have a large negative impact on pretraining on non-object-centric tasks where semantics and spatial relationships reveal essential information. This negative performance is restored through the removal of local crops and the addition of domain-specific augmentations. The same patterns hold for both datasets. An additional analysis using an alternate metric, the pointing game, used for interpretability assessment can be found in the appendix.
\vspace{-12pt}

\begin{figure}[!t]
    \begin{center}
    \centering
\includegraphics[width=0.9\linewidth]{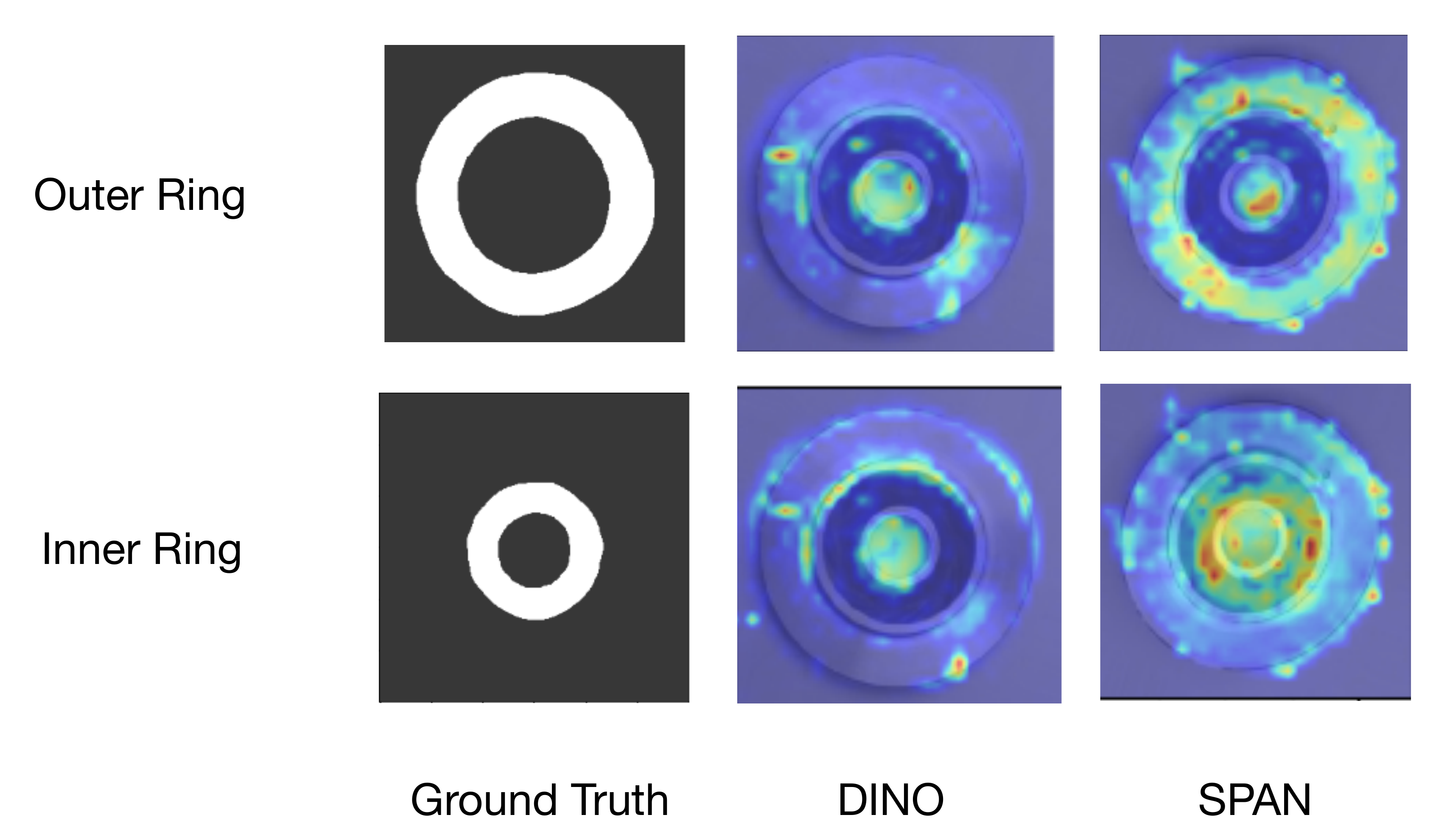}
    \caption{\textbf{Visualized QA attention maps from \ours and DINO.} After regularization, the attention maps from the attention heads with the highest IOU ground truth overlap were taken and visualized. The attention maps from domain-agnostic pretraining (DINO) are a lot more sparse and random. SPAN attention maps shows activation in regions of interest.}
    \label{fig:exQA}
    \vspace{-20pt}
    \end{center}
\end{figure}

\subsubsection{Qualitative Assessment}

In Figure~\ref{fig:MEPrune} and ~\ref{fig:exQA}, we visualize the attention maps resulting from the different models. Figure~\ref{fig:MEPrune} shows that DINO pretrained Vision Transformers are unable to learn salient representations from X-ray images. DINO even leads to collapse with worse representations than a DINO model not pretrained on chest X-rays at all, though removal of global-local crops and inclusion of medical imaging specific augmentations mitigates this performance drop. It also shows that \ours improves the alignment with the segmented regions and also disentangles the constituent attention maps across the heads. Similarly for manufacturing data, \ours improves and specializes the attention heads to focus on representations in the regions of interest compared to DINO.

\subsection{Downstream Disease Classification}

Classification performances are assessed using the mean receiver operating area under the curve (mAUC) score averaged over the 6 disease classes in the CheXpert classification challenge using a hold-out test set of 200 images: Atelectasis, Edema, Pleural Effusion, Cardiomegaly, Consolidation and No Finding. The linear classifier is trained on top of the frozen pretrained representations. The results in Table 2 show that \ours has a stronger multi-label classification performance compared to all baseline DINO variants. Specifically, \ours with the deformable image registration templates attains the highest mAUC score followed by the predicted global templates and the triangular spatial heuristics. This indicates that the more interpretable representations from \ours also lead to higher downstream performance as well. In line with the results of Section 4.2, DINO pretrained on ImageNet and CheXpert with domain-agnostic representations attains the lowest classification score. DINO only pretrained on ImageNet performs equally as DINO pretrained on ImageNet and CheXpert with domain-specific augmentations.

\input{tables/fig-B}

\subsection{Quality Assurance in Low Data Regimes}

The last set of experiments address the goal of data efficiency. The experiments measure the classification performances based on domain-agnostic (DINO) and \ours pretraining. Specifically, models are pretrained on the training set of 1,100 images. Models with the highest mAP overlap were selected to serve as the backbone to detect either defect or normal molds. They are then evaluated using a balanced set of samples with sizes of $\{4, 8, 16, 32\}$. The classification task setup is similar as described in the previous section where a linear probe is appended. The data were evaluated on a held-out test set of 200 images.

\begin{figure}[!t]
    \begin{center}
    \hspace{-1.5em}
    \centering
\includegraphics[width=0.8\linewidth]{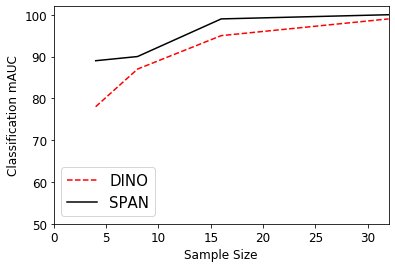}
    \caption{\textbf{Classification result of varying low-data regimes in QA.} The following plot shows the
test classification AUC with varying number of labeled examples for training in DINO and \ours. \ours outperforms DINO in all the chosen sample sizes. The largest performance increases are seen in the smaller sample sizes.}
    \label{fig:plotQA}
    \vspace{-20pt}
    \end{center}
\end{figure}

As visible in Figure ~\ref{fig:plotQA}, in the presence of only a handful of labels, \ours outperforms the DINO baseline by 11.2 AUC. However, as more samples are added to the training set, the model performances become more similar.

%% file: tables/fig-A.tex
\setlength{\tabcolsep}{0.11cm}
\begin{table}[t]
\begin{center}
\label{table:headings}
\begin{tabular}{|l|l|c|c|}
\hline
Part & Regimen & JSRT & Montgomery  \\
\hline \hline
Heart & DINO & 19.1 & - \\
  & DINO (CheX)  & 5.7 & - \\
  & DINO (CheX Augm) & 26.4 & - \\
  & \ours (Triangular) & 54.5 & - \\
  & \ours (Global Average) & 71.6 & - \\
  & \ours (Deformable) & \textbf{89.9} & - \\
\hline
 Left Lung & DINO & 30.1 & 43.2 \\
  & DINO (CheX)  & 22.0 & 16.5 \\
  & DINO (CheX Augm) & 25.1 & 40.7 \\
  & \ours (Triangular) & 59.2 & 40.0 \\
  & \ours (Global Average) & 71.3 & 84.1 \\
  & \ours (Deformable) & \textbf{88.3} & \textbf{90.0} \\
\hline
Right Lung & DINO & 46.3 & 35.0 \\
& DINO (CheX)  & 27.0 & 15.8 \\
& DINO (CheX Augm) & 36.5 & 27.6 \\
& \ours (Triangular) & 45.6 & 54.8 \\
& \ours (Global Average) & 82.5 & 50.3 \\
& \ours (Deformable) & \textbf{87.3} & \textbf{88.4} \\
 \hline
\end{tabular}
\vspace{5pt}
\caption{\textbf{Interpretability scores of attention heads.} The evaluation metrics included pixel-wise mAP on external validation sets where groundtruth segmentation masks were available. Due to the lack of heart segmentations in the Montgomery dataset, results of heart interpretability have not been reported. The results indicate that \ours improves the interpretability over DINO baselines.}
\end{center}
\vspace{-15pt}
\end{table}
\setlength{\tabcolsep}{1.4pt}

%% file: tables/fig-B.tex
\setlength{\tabcolsep}{4pt}
\begin{table}
\begin{center}
\label{table:headings}
\begin{tabular}{|l|c|}
\hline
Regimen & mAUC \\
\hline\hline
Random & 69.9 \\
DINO & 83.8 \\
DINO (Chexpert)  & 60.4 \\
DINO (Chexpert Augmentations) & 84.3 \\
\hline
\ours (Triangular) & 84.8 \\
\ours (Global Average) & 86.2\\
\ours (Deformable) & \textbf{86.5}\\
\hline
\end{tabular}
\end{center}
\caption{\textbf{Linear disease classification trained on a linear probe.} The pretrained models are used as feature extractors in the CheXpert classification task whereby a linear layer is fine-tuned to predict six diseases: Atelectasis, Pleural Effusion, Consolidation, Cardiomegaly, No Finding and Edema. The mAUC over all diseases are reported. \ours outperforms DINO pretraining methods for all different attention priors. DINO pretraining improves with domain-specific augmentations.}
\vspace{-10pt}
\end{table}
\setlength{\tabcolsep}{1.4pt}

%% file: 5-Conclusion.tex
\section{Conclusion}

We present \ours: a framework for knowledge-based self-supervised Vision Transformers, which incorporates useful inductive biases into the training processes that learn more interpretable representations, attain high evaluation scores in low-data regimes and lead to better performance on downstream classification tasks. Spatial Prior Attention (\ours) takes advantage of consistent spatial and semantic structure in unlabeled imaging datasets to guide Vision Transformer attention. Using chest X-ray radiographs and manufacturing data as primary case studies, we show that the resulting attention is even better when we combine \ours with the decreased data sample. The resulting attention masks are more interpretable than those from domain-agnostic pretraining, producing a 58.7 mAP improvement for lung and heart segmentation following the self-supervised pretraining. Additionally, \ours yields a 2.2 mAUC improvement compared to domain-agnostic pretraining when transferring the pretrained model to a downstream chest disease classification task. Lastly, we also show that \ours, at the lowest pretraining sample size of 4 (2 for each class), resulted in an increase of 11.2 mAUC in classification performance.

Our results indicate that the attention heads in self-supervised Vision Transformer can be specialized to attend to different objects and learn more semantic and meaningful representations underlying the data by embedding prior knowledge using our attention regularization framework.

%% file: 6-Appendix.tex
\begin{appendix}
\section{Appendix}
\subsection{Notations and Definitions}
\input{tables/fig-C}
\subsection{Data Augmentation Details}
DINO augmentations pretrained on chest images result in lower mAP overlap between the attention maps and ground truth segmentation masks, as observed in Table 1. To mitigate this, we use the radiograph-specific data augmentations proposed by Sowrirajan et al. \cite{sowrirajan2021moco}. These augmentations refrain from using local crops, color jittering and randomized grayscaling. Default augmentations do not apply well to gray-scale chest radiographs and are eliminated from the pretraining procedure only for the medical imaging tasks. Additionally, we also added rotations, translations, and random global crops, as well as adding brightness, sharpness and contrast. Each augmentation is applied with a random probability. When applied, the strength of the augmentation is sampled uniformly at random from the range of values specified in Table \ref{table:Datag}.

\input{tables/fig-E}

For the classification tasks, we do not apply any data augmentations. Pretrained self-supervised Vision Transformers are able to generalize well without augmentations when used as a feature extractor for linear classification evaluation. \cite{caron2021emerging} We only normalize the images prior to training.

\subsection{Training and Experiment Details}

The hyperparameters for training follow the defaults from DINO \cite{caron2021emerging} where possible. Pretraining is performed on the subset of frontal CheXpert images. Validation CheXpert images are used for saliency map evaluations. Additionally, JSRT and Montgomery validation images are used to score the interpretability of attention heads.
For pretraining, all images are rescaled to (224, 224). During inference, we rescale the image to (480, 480) to make sure the attention maps have a high enough resolution. All \ours models are pretrained for 30 epochs. The batch size is kept constant at 28. The default learning rate for DINO is $0.0005 * batchsize / 256$. However, we found that training \ours with a learning rate of 0.0001 created slightly better representations. $\lambda$ values are found via grid-search on a $log$-scale. The $\lambda$ values leading to the highest interpretability scores are chosen. 



\subsection{Additional Results}


\subsubsection{Interpretability}

The pointing game is a metric that measures the interpretability of attention maps compared to a ground truth segmentation or bounding boxes. \cite{zhang2018top} It evaluates whether the maximum entry of a saliency map falls within the region of interest, also called a hit. The metric is calculated as follows: $$Pointing Game = \dfrac{\# hits}{\# hits + \# misses}$$
Table \ref{table:pointing} shows that all \ours pretrained models outperform DINO by a large margin for both heart (80+\%) and lung (60+\%) interpretability. The results from the Montgomery dataset follow the same patterns from Table 1 where the increasingly specific templates in \ours allowed for better performance. This is generally the case for the JSRT results as well, but the results between the global average templates outperform the deformable model for the left lung and ties for the interpretability of the heart. Since there were no ground truth segmentation mask for the manufacturing dataset, we were not able to report results on it.
\input{tables/fig-D}

\begin{figure}[!t]
    \begin{center}
    \hspace{-1.5em}
    \centering
\includegraphics[width=0.5\textwidth]{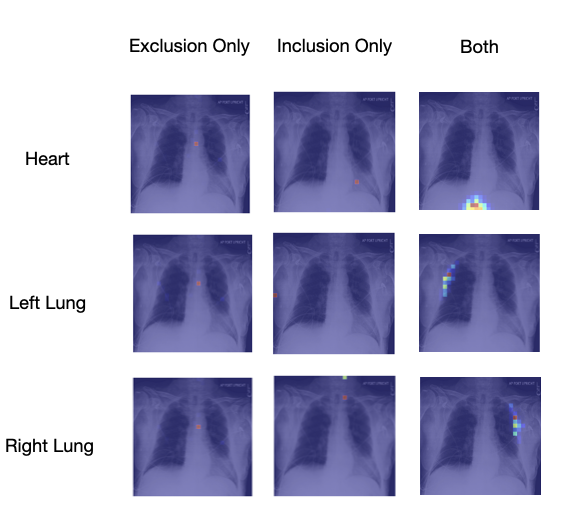}
    \caption{\textbf{Collapse in the absence of either regularization terms.} This graphic shows the effect of the absence of either regularization terms. As can be seen, all the mass of the attention maps resides in a limited area when only either regularization term is included.}
    \label{collapse}
    \end{center}
\end{figure}

\subsubsection{Regularization \& Collapse}

In this experiment, we investigate the effect of excluding the inclusion or the exclusion regularization loss terms by probing the attention maps and calculating the ground truth overlap between the created map and a ground truth segmentation for the heart, left lung and right lung using mAP. The attention map with the highest IOU is chosen for evaluation. The results in Table below show that in the absence of either terms the pretrained models perform worse. Closer inspection of these attention heads in Figure \ref{collapse} reveals empirically that pretraining collapses when either term is absent. This issue is mitigated when both loss terms are present.

\input{tables/fig-G}


\subsection{Self-Attention Visualizations}

\begin{figure*}[!t]
    \begin{center}
    \hspace{-1.5em}
    \centering
\includegraphics[width=1\textwidth]{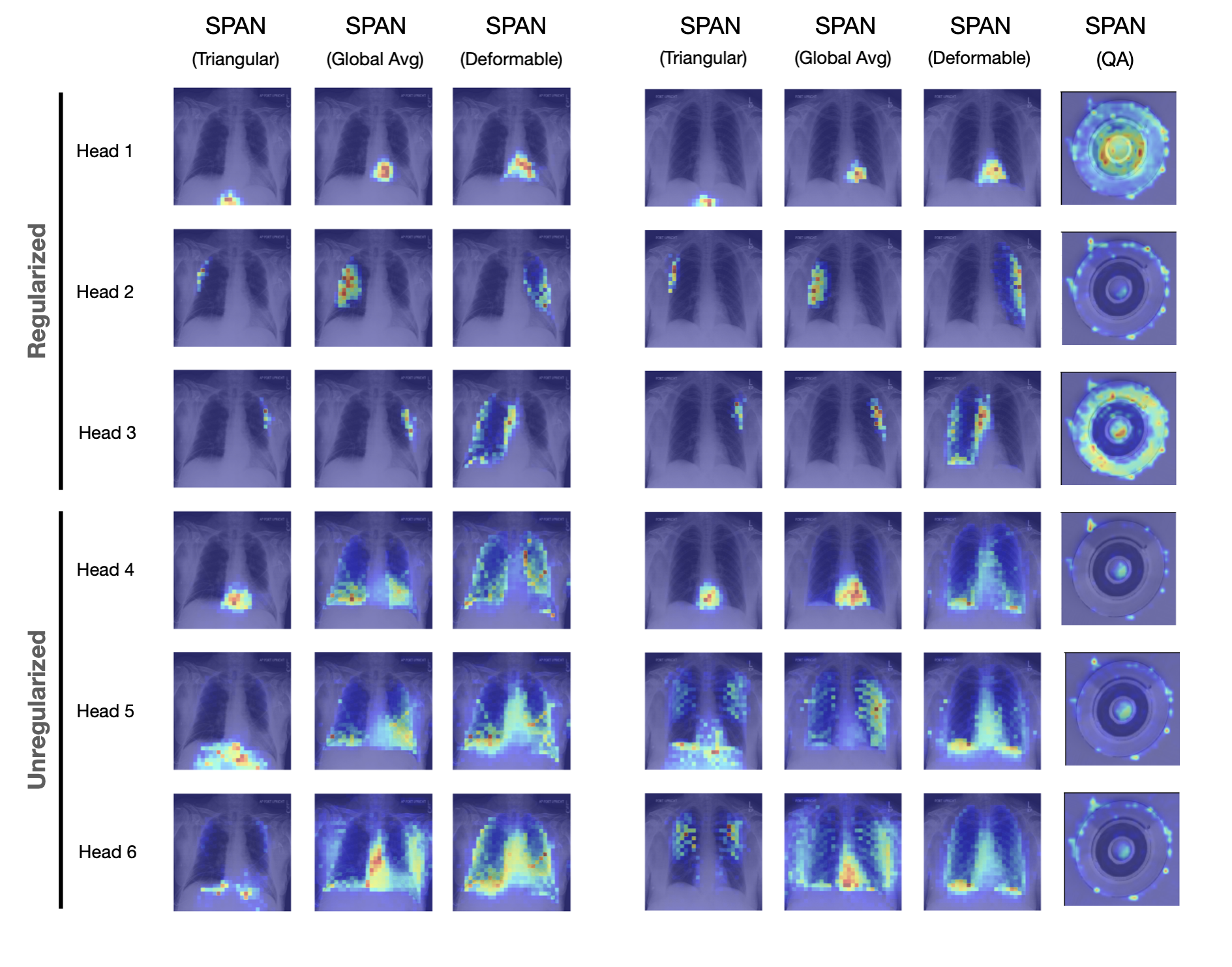}
    \caption{\textbf{Visualized self-attention over all heads.} This Figure shows the attention maps from \ours given samples from the CheXpert and Manufacturing QA dataset respectively. The regularized heads attend to regions and semantics, whereas the unregularized heads also learn emergent, interpretable representations. For instance, for the SPAN models - the first three heads consistently look at the designated areas (heart), left and right lungs. For the QA models, the heads were regularized to look at the black inner band, outer band and the middle band of the molds, which the model attends to successfully.}
    \label{fig:final}
    \end{center}
\end{figure*}

We sample two validation images from the CheXpert dataset and provide the attention maps over all 6 attention heads in Figure \ref{fig:final}. The Figure shows that the unregularized heads become more interpretable as the quality of the representations in the regularized heads increase. We even observe emergent properties in unregularized heads as a result of this.

\end{appendix}

%% file: tables/fig-C.tex
\setlength{\tabcolsep}{4pt}
\begin{center}
\label{table:headings}
\begin{tabular}{|l|c|}
\hline
Notation & Definition \\
\hline\hline
$m_{\theta_{s}}$ & Student model \\
\hline
$m_{\theta_{t}}$ & Teacher model \\
\hline
$X$ & Original image \\
\hline
$w$ & Image width\\
\hline
$h$ & Image height\\
\hline
$n_h$ & The number of attention heads \\
\hline
$n$ & The number of regularized\\
& attention heads \\
\hline
$p$ & Patch size\\
\hline
$P$ & The total number of patches \\
\hline
$V$ & The set of data augmentations to \\
 & apply to each image \\
 \hline
$\{X_1, ..., X_n\}$ & The set of augmented images \\
 &  acquired resulting from applying $V$ on $X$\\
 \hline
$X^{g}$ & Global crop \\
\hline
$P(x)$  & Output probability distribution\\
\hline
$\tau$ & Temperataure \\
\hline
$H$ & Cross-entropy \\
\hline
$A$ & Attention matrix \\
\hline
$A_s$ & Self-attention map \\
\hline
$K$ & Knowledge map \\
\hline
$\lambda_1$, $\lambda_2$ & Magnitude for inclusion \\ 
& and exclusion regularization \\
\hline
$X^{e}$ & Exemplar image\\
\hline
$X^{i}$ & Training image\\
\hline
$S^{e}$ & Ground-truth segmentation for exemplar\\
\hline
$S^{i}$ & Ground-truth segmentation \\ 
& for training sample\\
\hline
$\phi$ & Deformable transformation \\
\hline
\end{tabular}
\end{center}
\vspace{-10pt}
\setlength{\tabcolsep}{1.4pt}

%% file: tables/fig-E.tex
\setlength{\tabcolsep}{4pt}
\begin{center}
\label{table:headings}
\begin{tabular}{|l|c|c|}
\hline
Data augmentation & Probability & Range [min, max] \\
\hline \hline
Rotation & 0.2 & [-30, 30]  \\
Contrast & 0.1 & [0, 0.2]  \\
Brightness & 0.1 & [0, 0.2] \\
Sharpness & 0.1 & [0, 0.2]  \\
Random Crop & 1 & 224 x 224  \\
\hline
\end{tabular}
\label{table:Datag}
\end{center}

%% file: tables/fig-D.tex
\setlength{\tabcolsep}{0.11cm}
\begin{table}
\begin{center}
\caption{\textbf{Additional interpretability scores for the attention heads with highest overlap.} In this additional analysis, we included the \textbf{pointing game evaluation scores}, another metric for interpretability of attention maps. The figure shows that \ours outperforms all variants of DINO pre-training.}
\label{table:headings}

 
\begin{tabular}{|l|l|c|c|}
\hline
Part & Regimen & JSRT & Montgomery  \\
\hline \hline
Heart & DINO  & 9.2  &   \\
& DINO (CheX)  & 0 &  \\
  & DINO (CheX Augm)  & 10.7 &   \\
  & \ours (Triangular) & 62.0 &   \\
  & \ours (Global Average) & 98.0 & \\
  & \ours (Deformable) & \textbf{99.0} &  \\

 Left Lung & DINO & 26.9 & 57.97 \\
 & DINO (CheX)  & 0 & 0 \\
  & DINO (CheX Augm)  & 38.1 & 67.4 \\
  & \ours (Triangular) & 96.4 & 35.0 \\
  & \ours (Global Average) & \textbf{97.9} & 100 \\
  & \ours (Deformable) & 97.5 & \textbf{99.3} \\
 
 Right Lung & DINO & 65.9 & 39.3 \\
 & DINO (CheX)  & 0 & 0 \\
 & DINO (CheX Augm)  & 34.5 & 43.5 \\
 & \ours (Triangular) & 79.7 & 69.0 \\
  & \ours (Global Average) & \textbf{100.0} & 90.58 \\
  & \ours (Deformable) & \textbf{100.0} & \textbf{98.6} \\
 \hline
\end{tabular}
\label{table:pointing}
\end{center}
\end{table}

%% file: tables/fig-G.tex

\setlength{\tabcolsep}{4pt}
\begin{center}
\label{table:headings}
\begin{tabular}{|l|c|c|c|}
\hline
Regularization Term  & Heart & Left Lung & Right Lung\\
\hline
Exclusion Only & 34.7 & 21.1 & 13.0 \\
Inclusion Only & 14.9 & 52.0 & 41.1 \\
Both  & \textbf{54.5} & \textbf{59.2} & \textbf{45.6} \\
\hline
\end{tabular}
\label{reg}
\end{center}